\documentclass[conference]{IEEEtran}
\IEEEoverridecommandlockouts
\usepackage{cite}
\usepackage{amsmath,amssymb,amsfonts}
\usepackage{algorithmic}
\usepackage{graphicx}
\usepackage{textcomp}
\usepackage{xcolor}
\def\BibTeX{{\rm B\kern-.05em{\sc i\kern-.025em b}\kern-.08em
    T\kern-.1667em\lower.7ex\hbox{E}\kern-.125emX}}
\begin{document}

\title{Evaluating how LLM annotations represent diverse views on contentious topics}

\author{\IEEEauthorblockN{Megan A. Brown}
\IEEEauthorblockA{\textit{School of Information} \\
\textit{University of Michigan}\\
Ann Arbor, MI USA \\
mgnbrown@umich.edu}
\and
\IEEEauthorblockN{Shubham Atreja}
\IEEEauthorblockA{\textit{School of Information} \\
\textit{University of Michigan}\\
Ann Arbor, MI USA \\
satreja@umich.edu}
\and
\IEEEauthorblockN{Libby Hemphill}
\IEEEauthorblockA{\textit{School of Information} \\
\textit{University of Michigan}\\
Ann Arbor, MI USA \\
libbyh@umich.edu}
\and
\IEEEauthorblockN{Patrick Y. Wu}
\IEEEauthorblockA{\textit{Department of Computer Science} \\
\textit{American University}\\
Washington, D.C. USA \\
patrickwu@american.edu}
}

\maketitle

\begin{abstract}
Researchers have proposed the use of generative large language models (LLMs) to label data for research and applied settings. This literature emphasizes the improved performance of these models relative to other natural language models, noting that generative LLMs typically outperform other models and even humans across several metrics. Previous literature has examined bias across many applications and contexts, but less work has focused specifically on bias in generative LLMs' responses to subjective annotation tasks. This bias could result in labels applied by LLMs that disproportionately align with majority groups over a more diverse set of viewpoints. In this paper, we evaluate how LLMs represent diverse viewpoints on these contentious tasks. Across four annotation tasks on four datasets, we show that LLMs do not show systematic substantial disagreement with annotators on the basis of demographics. Rather, we find that multiple LLMs tend to be biased in the same directions on the same demographic categories within the same datasets. Moreover, the disagreement between human annotators on the labeling task---a measure of item difficulty---is far more predictive of LLM agreement with human annotators. We conclude with a discussion of the implications for researchers and practitioners using LLMs for automated data annotation tasks. Specifically, we emphasize that fairness evaluations must be contextual, model choice alone will not solve potential issues of bias, and item difficulty must be integrated into bias assessments.
\end{abstract}


\section{Introduction}
Recent research on large language models (LLMs) emphasizes their improved performance in a variety of natural language processing tasks, including classification and text generation \cite{tan2024large}. 
In the social sciences, researchers have proposed using generative LLMs to label texts for concepts of interest, usually in zero-shot settings \cite{rathje2024gpt,bail2024can,tornberg2023chatgpt,atreja2024prompt,wu2025semanticallyunrelated}.
While research shows that machine learning models frequently overrepresent the views of majority groups (see, e.g., \cite{davani2022dealing,prabhakaran2021releasing}), this has not been systematically studied for generative LLMs.
The core concern motivating our paper is that by using these models for text annotation in social science research, such research will over-represent the views of majority groups and under-represent the views of minority groups. 

To better understand this concern, we propose systematically evaluating the performance of a suite of generative LLMs on four different text annotation tasks. 
We use the NLPositionality dataset \cite{santy2023nlpositionality}, which contains labels on social media comments for toxicity; the POPQUORN offensiveness and politeness datasets \cite{pei2023annotator}, which contains labels on social media comments for offensiveness and labels on emails for politeness; and the Wikipedia comments dataset \cite{wulczyn2017ex}, which contains labels on comments from Wikipedia edits for toxicity. 
We selected these datasets because (1) the underlying tasks are subjective and often contentious in nature, resulting in disagreement among human raters; and (2) these datasets includes the demographics of annotators who provided the labels, allowing us to evaluate the performance of LLMs on these tasks and understand the extent to which LLMs agree with different demographic groups. 
We focus specifically on gender, race, and education because they were available in all the datasets; prior work also acknowledges the potential bias related to these groups (see, e.g., \cite{thakur2023unveiling,omar2025evaluating,kotek2023gender,kotek2024protected}). 

For LLMs, we use GPT-4o mini, Llama 3.3 70B Instruct, and Gemini 1.5 Flash. 
These are three of the most commonly used LLMs in social science research.
To ensure that our results are not prompt-specific, we also tested three semantically equivalent prompts for generating labels for each dataset. 
We label each text from each dataset using each of the three models and each of the three semantically equivalent prompts, resulting in nine LLM labels per text. 
See Table \ref{tab:experiment_design} for a summary of these factors and their levels. 
We determine agreement by evaluating whether the model generates the same label that the annotator did (e.g., if the annotator labeled the text toxic and the model labeled the text toxic, then the model agrees with that annotator). 
We compute the LLMs' agreements with each individual annotator and use logistic regression to understand the demographic factors that predict LLM agreement with human annotators. 

\begin{table}[!ht]
\centering
\caption{Factors and Levels in our Experiments}
\begin{tabular}{p{1cm}|p{1.45cm} p{1.45cm} p{1.4cm} p{1.4cm}}
\hline
Factor & \multicolumn{3}{l}{Factor Levels} \\
\hline
Models & GPT-4o mini & Llama 3.3 70B Instruct & Gemini 1.5 Flash \\
Datasets (tasks) & NLPositionality (toxicity) & POPQUORN Politeness (politeness) & POPQUORN Offensiveness (offensiveness)  & Wikipedia comments (toxicity) \\
LLM Annotations & 2,583 & 13,500 & 33,462 & 25,678 \\
Prompts & Prompt Version 1 & Prompt Version 2 & Prompt Version 3 \\
\hline
\end{tabular}
\label{tab:experiment_design}
\end{table}

Across all datasets, we find that LLMs do not show systematic and substantial disagreement with annotators on the basis of demographics.
While there are statistically significant biases, the direction of the bias of each demographic category is generally the same across all LLMs.
Additionally, when LLMs tend to agree with one demographic group over the other, the demographic group is not consistent across datasets. 
In other words, biases tend to be dataset-specific rather than LLM-specific.
For example, the LLMs tend to agree with white annotators more often on the offensiveness and politeness tasks, but tend to agree more with non-white annotators on the NLPositionality toxicity task. 
Within each task, there is no variation among the models on the demographic groups with which the models tend to agree: GPT 4o-mini, Llama, and Gemini agree with the same demographic groups. 
Moreover, the difficulty of the labeling task (measured using disagreement among human annotators) is much more predictive of LLM agreement than the demographics of the annotators.
Our findings indicate that LLM agreement with specific demographic groups may be a function of the data and how it interacts with the LLM's pretrained embedded information, rather than a function of systematic biases of the LLM across all subjective annotation tasks.

In summary, our contributions are as follows.
\begin{itemize}
    \item We provide the first systematic evaluation of demographic bias in LLMs when using LLMs for social science annotation tasks. We use four datasets, three popular LLMs, and three semantically equivalent prompts for each dataset in our analysis. Beyond analyzing demographic features, we also explicitly model item difficulty using entropy.
    \item We find that LLMs do not show systematic disagreement with annotators on demographics (specifically, gender, race, and education). We find that all LLMs tend to be biased in the same directions on the same dataset. Moreover, the difficulty of the labeling task is most predictive of LLM agreement with human annotators. 
    \item We discuss the implications for researchers and practitioners interested in using generative LLMs for automated data annotation tasks. Most importantly, we emphasize that fairness evaluations should be context-specific, model choice alone will not solve bias, and item difficulty must be integrated into bias assessments.
\end{itemize}

\section{Related Work}
\subsection{Bias in LLMs}
Previous research has focused on bias in natural language processing, highlighting how embeddings and language models can embed and manifest implicit biases existing in society \cite{bolukbasi2016man,caliskan2017semantics,garg2018word,zhao2019gender}.
Further, burgeoning research on LLMs shows that similar to their predecessors, LLMs exhibit biases related to gender, race, and other demographic factors on a range of tasks such as job recommendations, \cite{salinas2023unequal}, healthcare \cite{Omar2025}, and occupation prediction \cite{kotek2024protected}. 
Multiple studies have found that gender bias is prevalent in LLM outputs, manifesting in gendered word associations and biased narratives \cite{thakur2023unveiling,omar2025evaluating} as well as gendered stereotypes \cite{kotek2023gender}. 
Additional studies have also found racial biases \cite{omar2025evaluating}, such as medical LLMs generating racist stereotypes or biased outcomes \cite{hastings2024preventing}.
LLMs exhibit intersectional biases in addition to racial and gender biases, with LLM-generated texts implying lower levels of agency for multiply marginalized demographic groups (e.g., Black females) \cite{wan2024white}. 
LLMs also exhibit biases beyond just race and gender, including other protected groups such as sexuality and religion \cite{kotek2024protected}.
Studies have also found that LLMs exhibit sociodemographic biases even when they decline to respond \cite{tang2023llamasreallythinkrevealing}.
Moreover, LLMs may not only produce bias in the text they produce, but may express even greater bias when instructed to answer using personas \cite{dong2024can}. 
Our work builds on this literature by specifically looking at bias in subjective social science annotation tasks using generative LLMs. 

\subsection{Persona Effects in LLMs}
Persona effects are the ability of an LLM to embody a particular persona. 
For example, given a set of demographics, political beliefs, psychological profiles, or other information, LLMs can create responses that align with individuals of those demographics. 
LLMs demonstrate consistent ability to manifest assigned personas, but their impact varies widely across different tasks and contexts, with some configurations of identities significantly affecting performance and others showing minimal influence. 
The authors of \cite{hu2024quantifying} found that personas account for less than 10\% of the variance in data annotations, but including these annotations in LLM prompts significantly improves annotation accuracy; in other words, persona assignment can shape LLM behaviors in measurable ways.
However, these benefits are not universal. 
Prompting job-related identities did not show measurable improvements \cite{zheng2024helpful}.
Persona assignment can also produce overly specific or stereotyped results. 
The authors of \cite{gupta2023bias} showed that persona assignment used stereotypes when responding to questions after being assigned a persona, and the authors of \cite{bisbee2024synthetic} showed that persona assignment to produce political opinions yields results with significantly less variation than the underlying population the LLM is attempting to represent. 
Taken together, the existing research on persona effects in LLMs shows that LLMs can accurately represent the viewpoints given a set of personas, but these viewpoints can also be biased with respect to the personas the researchers are aiming to represent.
Our work does not use prompts that induce a persona. We use three semantically equivalent prompts that do not contain any demographic information. Our aim is to better understand if the LLM's generated response to identity-free prompts more strongly correlates with specific demographic categories.

\subsection{LLMs for Text Labeling}
There is a growing body of research evaluating the efficacy of generative LLMs for text labeling tasks (see, e.g., \cite{tan2024large,atreja2024prompt}). 
Many of these studies find that LLMs exhibit higher performance in classification tasks, and LLMs can be cheaper than hiring human annotators \cite{gilardi2023chatgptoutperforms,li2023coannotating,tornberg2024llmsexperts}. 
Moreover, there are particular tasks where LLM labeling can yield pro-social results, such as using LLMs to label toxic online speech rather than exposing moderators to harmful online content \cite{li2024hot}. 
Among subjective annotation tasks, which are often the subject of papers proposing the use of LLMs (e.g., \cite{tornberg2024llmsexperts}), it is unclear whether LLMs are biased in their annotations. Our paper aims to analyze whether LLMs are biased in their generated labels on subjective annotation tasks.

\subsection{Annotators and Subjective Annotation Tasks}
Models are frequently trained on data labeled by annotators.
For subjective annotation tasks---tasks where the label may not have a ground truth and instead rely on the subjective judgments of the data annotators---data annotations can replicate the same biases that exist in humans \cite{hube2019understanding,diaz2020biases,bavaresco2025llmsinsteadhumanjudges}. 
Recent research on subjective data annotations shows that annotator characteristics influence task outcomes \cite{al2020identifying,hube2019understanding,diaz2020biases}. 
Aggregating over annotators to determine a ground truth label (for example, by taking the majority label across all annotators) can systematically underrepresent the views of minority annotators \cite{davani2022dealing,prabhakaran2021releasing}. 
The pool of annotators---their demographics, cultural backgrounds, political beliefs, and personality traits---and the aggregation decisions made by researchers can create biased training data.
Machine learning models are then trained on these biased annotations, manifesting bias in model outputs. 
\cite{diaz2020biases} shows how annotator bias in subjective annotation tasks biases downstream model predictions. 

Disagreement in annotations is not necessarily a result of poor performance on the task, but can be the consequence of different subjective beliefs \cite{wan2023everyone}.
Using LLMs to label texts essentially treats an LLM as a data annotator, which may suffer from the same biases exhibited when aggregating over multiple annotators since LLMs can be understood as aggregations over large amounts of training data. We aim to analyze this type of bias that LLMs may exhibit on subjective annotation tasks.

To combat this, some works have attempted to build into their classification pipelines a diverse set of perspectives, often called perspective-aware models (see, e.g., \cite{casola-etal-2023-confidence,orlikowski-etal-2023-ecological,casola-etal-2024-multipico}). However, our work focuses exclusively on using generative LLMs without any additional gradient-based updates or architectural modifications to the models. Our prompt-based approach largely aligns with how social scientists typically use these models.

The work closest to ours is \cite{sun-etal-2025-sociodemographic}. They also analyzed the POPQUORN dataset's offensiveness and politeness datasets to understand how LLM annotations may differentially correlate with specific demographic categories. Their findings on this dataset are similar to ours. In our work, however, we looked at two additional datasets, allowing us to better understand if the same patterns generalize outside the POPQUORN suite of datasets. We also explicitly modeled text classification difficulty using an entropy approach, while they used random effects for each item. 

\section{Data \& Methods}
To account for the diversity of language tasks researchers may use LLMs for, we used four open datasets covering four annotation tasks in this study \cite{santy2023nlpositionality,pei2023annotator,wulczyn2017ex}. 
These datasets are labeled for contentious topics, which are the subject of much research related to LLM annotations due to the potential for reducing human annotators' exposure to harmful content \cite{li2024hot}. 
These datasets are also among the few in NLP that include the demographics of annotators, allowing us to understand how LLM annotations represent (or do not represent) the views of different annotators. 
We describe the datasets below and outline our analytical method for evaluating LLM annotation biases. 
Table~\ref{tab:demographicsummary} shows a summary of the demographics in each dataset. 
For education, the majority group is non-college educated and the minority group is college educated; for gender, the majority group is men and the minority group is non-men; and for race, the majority group is white and the minority group is non-white. 

\subsection{NLPositionality Toxicity Data}
The NLPositionality Toxicity Data is a collection of social media posts that contain binary labels indicating whether or not those posts are toxic \cite{santy2023nlpositionality}.
Annotators were sourced through Lab in the Wild, an opt-in crowd-sourcing platform to evaluate the alignment of models with different demographic groups. 
From this dataset, we used gender, race, and education demographics. 
There are 299 unique texts annotated by 1,082 unique annotators.

\subsection{POPQUORN Offensiveness Data}
The POPQUORN Offensiveness dataset contains social media comments with ordinal labels on a scale of 1 to 5 (1 being not offensive at all and 5 being very offensive).
The annotators are sourced through a quota-sampled pool of workers on Lucid with the aim of being generally representative of the U.S. population. 
From this dataset, we used gender, race, and education demographics. 
The POPQUORN Offensiveness dataset consists of 1,500 unique comments randomly sampled from the Ruddit dataset \cite{hada2021ruddit}.

\subsection{POPQUORN Politeness Data}
The POPQUORN Politeness data set contains emails with ordinal labels on a scale of 1 to 5 (1 being not polite at all and 5 being very polite).
The annotators are sourced through a quota-sampled pool of workers on Lucid with the aim of being generally representative of the U.S. population, though from a different pool than the POPQUORN Offensiveness dataset. 
From this dataset, we used gender, race, and education demographics. 
The dataset contains 3,718 unique emails from \cite{shetty2004enron}. 

\subsection{Wikipedia Toxicity Data}
The Wikipedia toxicity data set contains comments in Wikipedia edits which contains binary labels for comments indicating whether or not they are toxic \cite{wulczyn2017ex}. 
The annotators were sourced on CrowdFlower. 
From this dataset, we used gender and education demographics; this dataset does not contain information about annotators' race. 
The dataset consists of 159,463 unique texts.
Due to budget constraints, we used stratified samples of this dataset.
We stratified this sample based on label entropy, ensuring that our sample is balanced across texts of varying levels of disagreement among annotators. 
This sampling process yielded 2,095 unique texts. 

% latex table generated in R 4.4.2 by xtable 1.8-4 package
% Fri Mar 28 17:42:25 2025
\begin{table}[!ht]
\centering
\caption{Demographics of annotators across datasets. POPQUORN is abbreviated as PQ.}
\label{tab:demographicsummary}
\begin{tabular}{llccc}
  \hline
Dataset & Demographic & Minority & Majority & NA \\ 
  \hline
NLPositionality & Education & 614 & 253 & 214 \\ 
NLPositionality & Gender & 711 & 336 &  34 \\ 
NLPositionality & Race/Ethnicity & 766 & 296 &  19 \\ 
PQ Offensivness & Education & 171 &  85 &   6 \\ 
PQ Offensivness & Gender & 138 & 124 &   0 \\ 
PQ Offensivness & Race/Ethnicity &  65 & 197 &   0 \\ 
PQ Politeness & Education & 334 & 155 &  17 \\ 
PQ Politeness & Gender & 267 & 236 &   3 \\ 
PQ Politeness & Race/Ethnicity & 135 & 367 &   4 \\ 
Wikipedia & Education & 2,315 & 1,039 &   2 \\ 
Wikipedia & Gender & 1,161 & 2,195 &   0 \\ 
   \hline
\end{tabular}
\end{table}

\subsection{Data Processing and Sampling}

To assess whether LLM models are more likely to align with socially dominant demographic groups, we recoded gender, race, and education variables into binary categories: man/non-man, white/non-white, and college-educated/non-college-educated.
This transformation was undertaken to make the datasets comparable, as each dataset had a different breakdown of each demographic.
Such a transformation also allowed us to analyze agreement patterns without focusing on the demographic predictors themselves, instead highlighting potential biases favoring dominant social groups.

For each text, we calculated the normalized label entropy by taking the set of human labels for a given text, calculating entropy, and dividing by the log number of labels. More specifically, 
\[\text{Label Entropy} = \frac{-\sum_{x}^{n} p(x) \log p(x)}{\log(n)}\]
where $p(x)$ represents the probability of an annotator selecting a particular label and $n$ represents the total number of labels. 
We used label entropy as a measure of disagreement among human annotators. 
A low label entropy indicated a high level of agreement across annotators; a high label entropy indicated a low level of agreement among annotators. 
Disagreement among annotators can be a result of task difficulty, where texts that do not neatly fit into one category or another have high levels of disagreement due to the difficulty of discernment. 
However, high label entropy can also indicate legitimate disagreement because of subjective views of the concept of interest. 

\subsection{LLM Annotations}
For the experiments, we employed GPT-4o Mini, Llama 3.3 70B Instruct, and Gemini 1.5 Flash. These models represent a diverse range of architectures, training processes, and openness, and are widely used in LLM research, enabling a broader analysis of model behavior across different LLMs researchers commonly use.
To mitigate prompt-specific biases, we designed three distinct but semantically equivalent prompts for each dataset.
These prompts can be found in the Appendix.
Each language model was prompted using each semantically equivalent prompt, generating labels for each input text accordingly.\footnote{For the LLM annotations, we operationalized ``toxicity'' as our annotation framework to bridge the methodological gap between the narrow ``hate speech'' criteria used by human coders and the broader ``rude, hateful, aggressive, or unreasonable language'' criteria applied to LLMs in the original NLPositionality study.} 
Utilizing three language models and three prompts, each text received a total of nine independent LLM-generated annotations, resulting in a total of 75,223 annotations across all four tasks. 
For each of the LLM annotations, we extracted the label from the response using GPT-4o mini (e.g. for the toxicity task, extracting ``yes'' from ``yes, the text is toxic.'')
We then compared the LLM annotations with human annotations, specifically evaluating whether the models' annotations aligned with the human annotations.
All experiments were run using OpenAI's API for GPT-4o mini, DeepInfra for Llama 3.3 70B Instruct, and the Google Gemini Developer API for Gemini 1.5 Flash. We used default hyperparameters (temperature and top-$p$) for all models.

To examine factors influencing agreement between LLM and human annotations, we used logistic regression models. 
Specifically, our dependent variable was model-human agreement, where agreement is 1 and disagreement is 0, and our independent variables were the demographic attributes of the human annotators.
We also controlled for label entropy---which captures disagreement among human annotators---and prompt variation to account for any prompt-specific or data-specific effects on model responses. In other words, our logistic regression was the following.
\begin{align*}
    \text{Agreement} &= \beta_0 + \beta_{entropy} \text{Label Entropy} \\
    &+ \beta_{gender}\text{Gender(ref=man)} \\
    &+ \beta_{race}\text{Race(ref=White)} \\
    &+ \beta_{edu}\text{Education(ref=college educated)} \\
    &+ \beta_{prompt}\text{Prompt(ref=Prompt Version 3)}
\end{align*}
To account for the number of regressions we ran, we applied Bonferroni correction for significance testing.
For the main results, we performed a logistic regression model for each dataset and language model. 

\section{Results}
\begin{figure*}[!ht]
\centering
\includegraphics[width=0.8\linewidth]{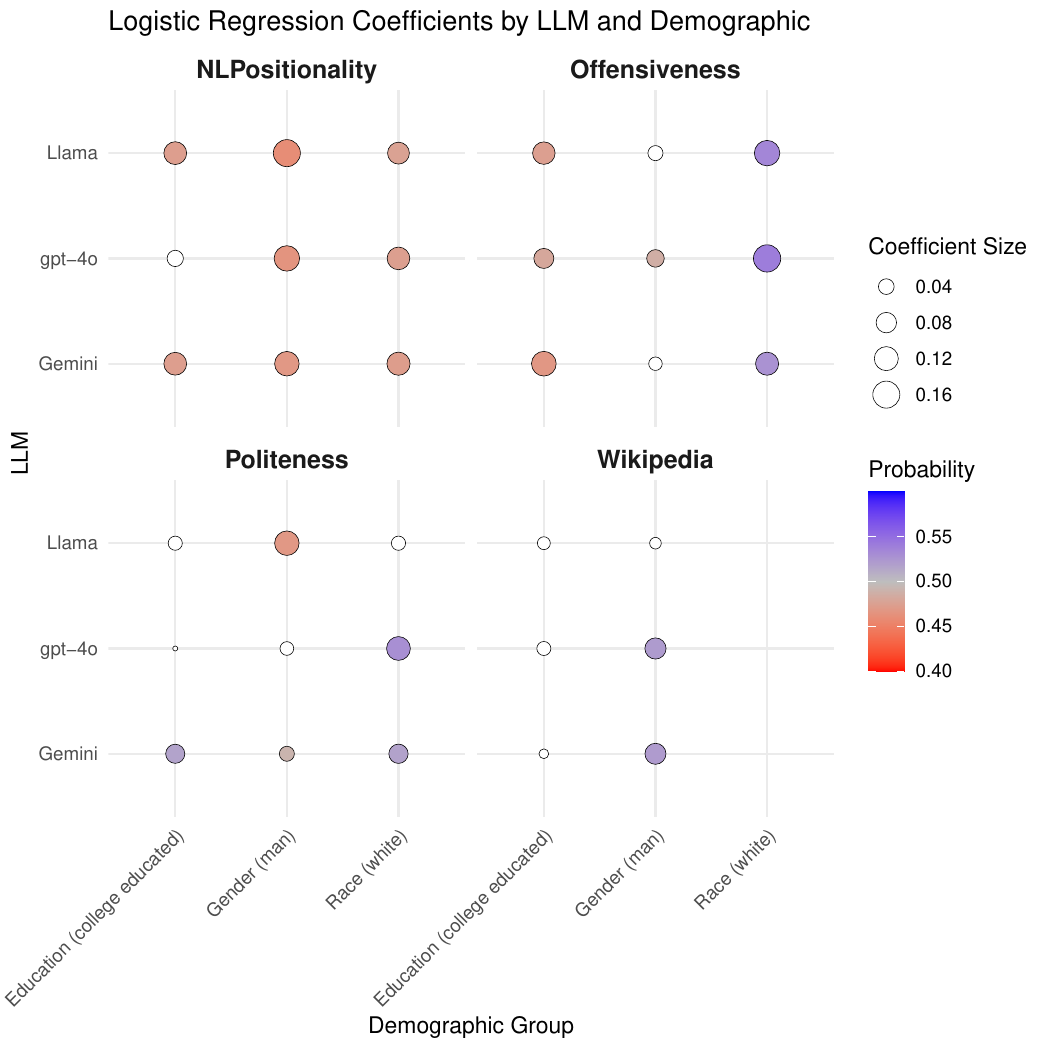}
\caption{\textbf{Logistic Regression Coefficients by LLM and Demographic Group}: The figure shows probabilities from logistic regressions for each LLM-dataset pair. Each facet represents one of the datasets. Within each facet, there is a grid representing regression coefficients for each demographic (columns) and LLM (rows). The point is sized by the coefficient size, and the point color is determined by the probability of agreement with the reference group (in parentheses). Blue indicates that the LLM is more likely to agree with the reference group, and red indicates that the LLM is more likely to agree with the non-reference group. White dots represent coefficients that were not statistically significant.}
\label{fig:regdemographics}
\end{figure*}

Figure~\ref{fig:regdemographics} shows the logistic regression results for each data set and LLM.
We present both the logistic regression coefficients and their corresponding probabilities of LLM agreement with each demographic group; the latter are derived from the regression coefficients and provide more intuitive interpretation of the results.
Colored points indicate statistically significant coefficients with $\alpha < 0.05$ (after Bonferroni corrections); blank points indicate non-significant coefficients.
A blue-colored point indicates that the LLM's generated responses agreed with the socially dominant demographic group (i.e., the reference category); a red point means that the LLM's generated responses agreed with the less socially dominant demographic group. 
The size of the point denotes the coefficient's magnitude. 

In the NLPositionality dataset, all LLMs were less likely to agree with college-educated annotators, annotators who identify as men, and white annotators. 
Within the POPQUORN offensiveness dataset, all LLMs were less likely to agree with college-educated annotators and more likely to agree with white annotators. 
GPT-4o mini was less likely to agree with annotators identifying as men, but Llama and Gemini did not exhibit a statistically significant likelihood of agreement with annotators who identify as men. 
For the POPQUORN politeness dataset, the LLMs were more likely to agree with college-educated annotators, less likely to agree with annotators who identify as men, and more likely to agree with white annotators, though some of these were not statistically significant. 
Lastly, for Wikipedia, the LLMs agreed with annotators who identify as men, and results for education were insignificant. 

Taken together, the way LLMs' generated responses agreed with annotators with specific demographics were generally consistent \textit{within} a dataset, but were \textit{not} model-specific. 
In other words, we found that within datasets, different LLMs were generally consistent regarding the demographic group they agreed with, indicating that agreement with different demographic groups was dataset-specific.
This consistency suggests that the task or domain drives the bias more than the choice of LLM.
We also found that the range of probabilities was between 0.46 and 0.55, indicating that even at their most biased, LLMs were only about 5\% more likely to agree with one demographic group over another. 
These different results across datasets suggest that LLMs do not exhibit a general tendency to systematically over-represent or under-represent the views of particular demographic groups, but rather may do so depending on the specific annotation task at hand.

\begin{figure*}[!ht]
\centering
\includegraphics[width=0.9\linewidth]{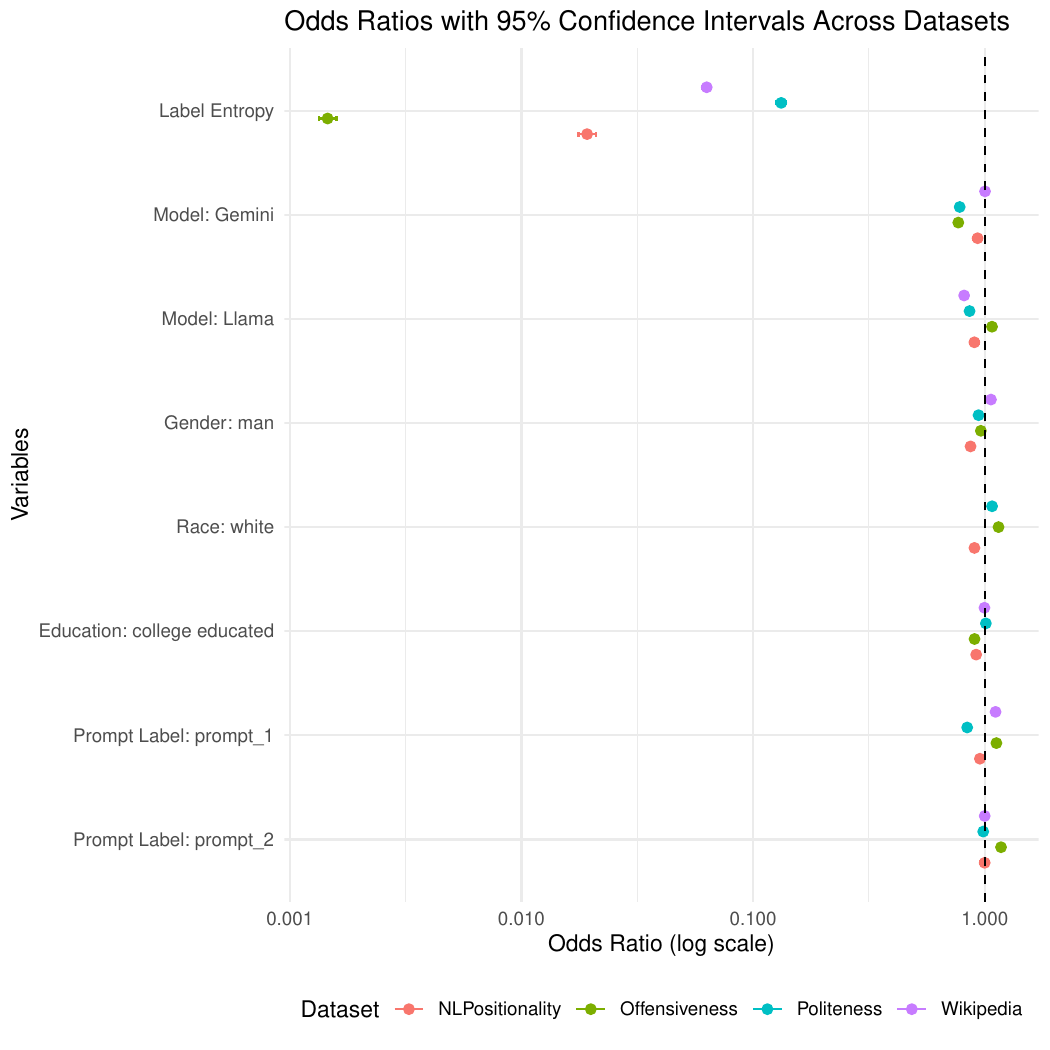}
\caption{\textbf{Odds Ratios Across Datasets}: The figure shows the odds ratios from logistic regressions for each dataset predicting LLM agreement as a function of label entropy, LLM, annotator gender, annotator race, annotator education, and prompt. The odds ratios are displayed with their 95\% confidence intervals. The x-axis is the odds ratio, and the y-axis is the variable of interest. Points are colored for each dataset: NLPositionality (red), POPQUORN Offensiveness (green), POPQUORN Politeness (blue), and Wikipedia comments (purple).}
\label{fig:oddsratios}
\end{figure*}

To further examine the factors contributing to LLM agreement with annotators, we turned to another logistic regression that used the same set of predictors as before along a categorical variable that coded the LLM used, allowing us to understand the influence of LLM choice on agreement.
Figure~\ref{fig:oddsratios} shows the odds ratios for each predictor for each dataset. 
As the figure shows, the strongest predictor for LLM agreement, by far, was label entropy. 
LLM agreement was inversely related to label entropy---as label entropy increased, the LLMs were less likely to agree with annotators. This intuitively makes sense: if annotators largely disagree with each other, then there are by default fewer possible annotators for the LLM to agree with. 
Additionally, texts with higher levels of disagreement can be considered texts trickier or more difficult to label, hence causing disagreement among the annotators and reducing the likelihood that the LLM would agree with any particular annotator.

To a smaller and less consistent degree, the LLM choice played a role in the agreement predictions.
For the NLPositionality and Politeness datasets, Llama and Gemini led to lower agreement than GPT-4o mini.
For the POPQUORN offensiveness dataset, Gemini was less likely to produce labels that agreed with annotators than GPT-4o mini, while Llama produced labels with higher rates of agreement with annotators. 
For Wikipedia, Llama was less likely to agree with annotators than GPT-4o mini. 
In short, the choice of the model did generally matter, but, consistent with our findings in our previous logistic regressions, there was no model that demonstrated agreement with annotators more consistently across all datasets. 

The semantically equivalent prompt used also mattered, but not consistently. Overall, the impact of the chosen prompt was lower than that of label entropy and the LLM used.

In short, label entropy was by far the most predictive factor of agreement between LLMs and human annotators. While there were significant odds ratios with model choice and prompt choice, the effect sizes were by far smaller than the effect sizes of label entropy. We note that we estimated a model without entropy to assess whether it was masking demographic effects. Removing entropy from the model did not meaningfully change results. Details are in the Appendix.

\section{Discussion}
We compared three LLMs' annotations on four tasks and did not find clear evidence of systematic bias toward any particular gender, race, or educational attainment demographic group.
Instead, we found that the bias was dataset-specific. All LLMs tended to be biased in the same direction on each dataset.
This bias was towards socially dominant demographic groups on some datasets, and in other datasets, the models were biased towards less socially dominant demographic groups. 
We also found that in cases where there was a statistically significant probability of an LLM agreeing with one demographic group relative to another, the probability of agreement was relatively small: the probabilities of agreeing with any demographic group were between 0.46 and 0.55. 
We found, instead, that the difficulty of the task was a much more significant predictor of the LLM agreement with human annotators.  

The core concern motivating our paper is that known biases in LLMs \cite{kotek2023gender,kotek2024protected,omar2025evaluating,hastings2024preventing,thakur2023unveiling} may manifest themselves in privileging the views of some demographic groups over others in data annotations, impacting the statistical inferences that use those annotations. 
Our findings suggest that these concerns are present, as demonstrated by the statistically significant coefficients in our logistic regressions, but they manifest in a dataset-specific manner.
The patterns are not indicative of a generalized, systematic bias in the models towards a particular set of demographic groups.
These findings highlight the nuance of LLM bias. 
It can be contrary to what we might expect from existing research on LLM bias, which demonstrates that LLMs embody and perpetuate stereotypes in their linguistic outputs \cite{kotek2024protected}. 
The lack of consistency in demographic agreement across datasets implies that any observed differences are more likely a function of the data characteristics and how it interacts with the LLM's pretrained embedded information rather than inherent and systematic biases embedded in the LLMs that manifest in all subjective annotation tasks.
This finding may also be related to recent findings that representations in AI models are converging, but explorations of this connection are outside the scope of this paper (see, e.g., \cite{huh2024prh}).

\subsection{Implications for Researchers and Practitioners}
For researchers and practitioners considering using generative LLMs for data annotation, our results indicate that while LLMs do not appear to introduce strong systematic demographic biases, challenges remain in handling ambiguous or highly subjective labeling tasks. We highlight four implications for researchers.

\paragraph{Fairness evaluations must be contextual} A specific LLM can appear pro-minority in one setting and pro-majority in another. The specific demographic group with which the LLM most strongly agrees, whether that is an annotator of a specific race, gender, demographic, or some combination of the three, may also vary by data set. We suggest that practitioners use a demographically balanced evaluation dataset to identify potential biases of the generative LLM used.

\paragraph{Model choice alone will not solve bias} As our findings indicate, LLMs tend to be biased in the same directions within the same dataset. Turning to a different LLM with a different architecture or training strategy may not effectively mitigate biases.

\paragraph{Persona-specific prompting may yield diminishing returns} The literature on persona effects \cite{hu2024quantifying} suggests persona-specific prompting can effectively increase agreement for particular demographic groups. However, our results indicate that there are diminishing returns to persona-specific prompting because the difference between demographic groups and LLMs' agreement is already relatively low. These small gains in accuracy are likely not worth the potential costs of persona-specific stereotyping those prompts can elicit. In some cases, the benefits of using LLMs to annotate data may outweigh representational harms. For example, it may be better to use LLMs to annotate data than relying on human annotators because the annotation task itself can cause significant harm to annotators. Exposing annotators to hate speech or toxic content carries risks for those annotators \cite{li2024hot,schopke2024volunteer}.

\paragraph{Effect sizes of bias are often significant but small, and item difficulty is usually the much stronger predictor of human disagreement with LLMs} It is important to balance the evaluation of biases with the fact that some texts, especially in subjective annotation tasks, are simply more difficult to label. In our results, the most significant predictor of LLM disagreement with annotators was label entropy, or disagreement among human annotators, indicating that LLMs and human annotators struggled with similar annotation tasks. It is critical to explicitly model item difficulty when examining potential demographic biases with the LLM's generated labels.

\subsection{Limitations and Future Work}
We are only looking at four datasets, and given the heterogeneity of our findings, we are unable to provide clear evidence regarding systematic, consistent biases in some of the most popular LLMs on subjective annotation tasks. 
Moreover, we only test three LLMs, so other LLMs may not behave this way or actually may be rather biased. 
We also only test datasets in English, and our results may not hold in other language contexts.
Additionally, for the closed source models we tested (GPT-4o mini and Gemini), models may be updated, meaning results may not be consistent over time. 
While open weight models may also change over time, researchers could use version control to use older versions of models \cite{munger2023temporal}.
In recognition of the potential harms of LLMs, LLM developers also use red-teaming and safety alignment techniques to regulate model outputs, which may also explain some of the results in this paper. 
However, due to lack of transparency from these developers about how these models were trained and what data was used, we cannot provide clear evidence on the effects of red-teaming and safety alignment on our results.
Future work should explore whether these findings hold across additional datasets and tasks, as well as investigate strategies for improving LLM performance in cases of high annotator disagreement. 
Additionally, while LLMs may not systematically favor particular demographic groups, their responses are still influenced by model choice and prompting strategy, underscoring the need for careful prompt engineering and validation when using these models for annotation in research and applied settings.

\section{Conclusion}
Our study evaluated whether large language models (LLMs) exhibit systematic bias in agreement with annotators from different demographic groups. 
Across four annotation tasks using four datasets and three LLMs, we found that LLM agreement with annotators was not systematically driven by specific demographic differences. 
Instead, LLMs tended to exhibit biases in the same directions within each dataset; in other words, the biases exhibited are not model-specific, but dataset-specific.
Moreover, by far the strongest predictor of LLM agreement was label entropy, or the level of disagreement among human annotators on the original annotation task. 
This suggests that rather than consistently favoring particular demographic groups, LLMs struggle most with annotating texts that human annotators themselves find ambiguous or contentious. 
Additionally, we found that the choice of LLM and the specific prompt used to elicit labels also influenced agreement rates, but no single model or prompt consistently outperformed the others across all tasks.
Practically, our results suggest that fairness evaluations must be contextual, model choice alone will not solve bias, and item difficulty must be integrated into bias assessments.

\bibliographystyle{IEEEtran}
\bibliography{bib}

\appendix
\section{Appendix}

\subsection{Robustness Checks}
\begin{figure}[!ht]
\centering
\includegraphics[width=\linewidth]{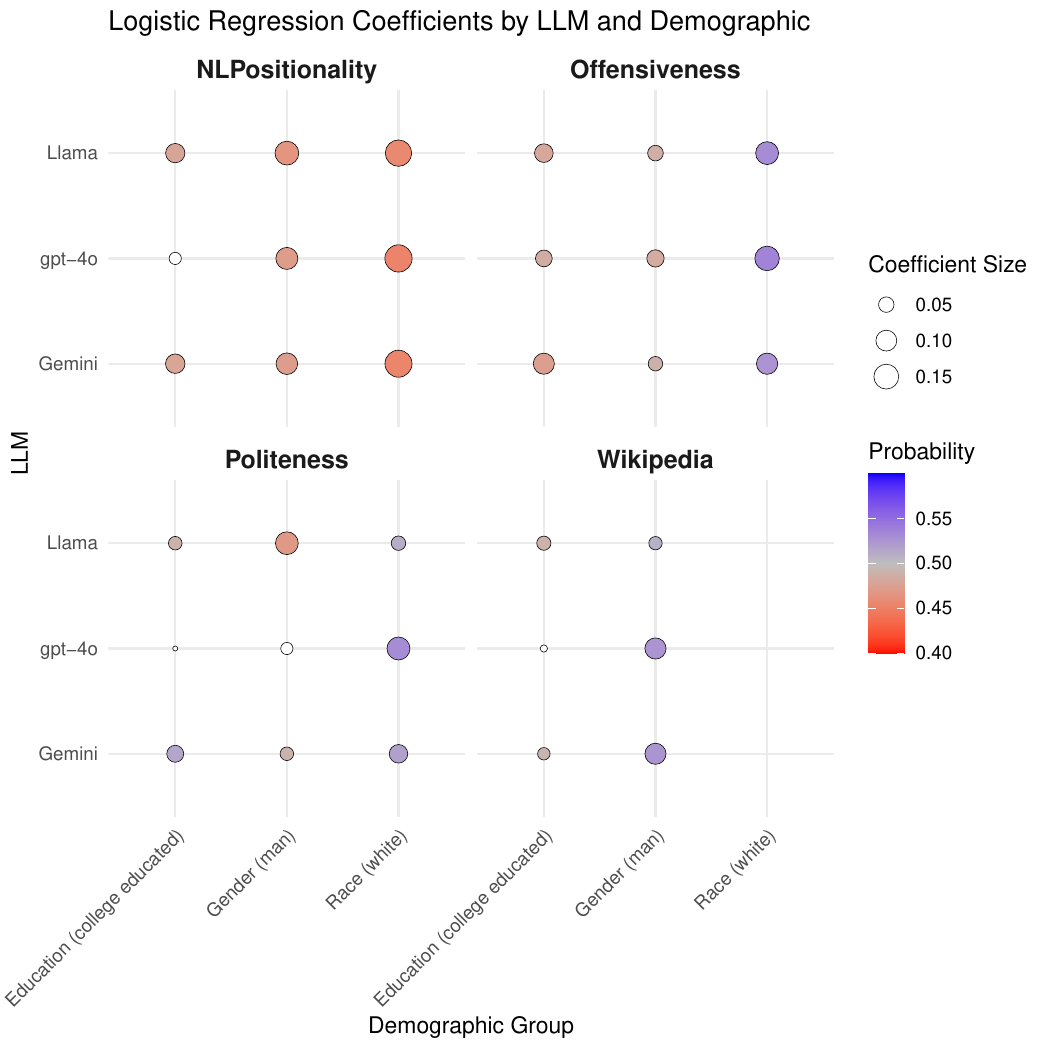}
\caption{\textbf{Logistic Regression Coefficients by LLM and Demographic Group}: The figure shows probabilities from logistic regressions for each LLM-dataset pair. Each facet represents one of the datasets. Within each facet, there is a grid representing regression coefficients for each demographic (columns) and LLM (rows). The point is sized by the coefficient size, and the point color is determined by the probability of agreement with the reference group (in parentheses). Blue indicates that the LLM is more likely to agree with the reference group, and red indicates that the LLM is more likely to agree with the non-reference group. White dots represent coefficients that were not statistically significant.}
\label{fig:regdemographicsappendix}
\end{figure}

To ensure that label entropy is not masking demographic variation, we run the same logistic regressions as specified in Figure~\ref{fig:regdemographics}, but we leave label entropy out of the controls. 
The results remain largely the same as our main specification, alleviating concerns that controlling for label entropy is masking true effects. The results can be found in Figure~\ref{fig:regdemographicsappendix}.

\subsection{Classification Prompts}

We used three semantically equivalent classification prompts with each LLM.

\textbf{NLPositionality Toxicity Data}
\begin{itemize}
    \item \textbf{Prompt 1}: I want you to perform a data annotation task. Your task is to carefully read the text provided below and determine whether the text is toxic or not. In your output, I want you to respond with yes if the text is toxic, otherwise respond with no. 
    \item \textbf{Prompt 2}: Perform data annotation by determining if the text is toxic or not. Only respond with yes or no.
    \item \textbf{Prompt 3}: Carefully review the text below and assess whether it is toxic. If the text is toxic, respond with yes; if not, respond with no.
\end{itemize}

\textbf{POPQUORN Offensiveness Data}
\begin{itemize}
    \item \textbf{Prompt 1}: I want you to perform an annotation task. Your task is to carefully read the text provided below and determine how offensive the text is on a scale of 1-5, where ``1'' is ``not offensive at all'' and 5 is ``very offensive.''
    \item \textbf{Prompt 2}: Complete an annotation task by carefully reading the text below and rating its offensiveness on a scale of 1 to 5, where ``1'' means ``not offensive at all'' and ``5'' indicates ``very offensive.'' 
    \item \textbf{Prompt 3}: Rate the offensiveness of the text below from 1 (not offensive) to 5 (very offensive).
\end{itemize}

\textbf{POPQUORN Politeness Data}
\begin{itemize}
    \item \textbf{Prompt 1}: I want you to perform an annotation task. Your task is to carefully read the email provided below and determine how polite the email is on a scale of 1-5, where ``1'' is ``not polite at all'' and 5 is ``very polite.''
    \item \textbf{Prompt 2}: Complete an annotation task by carefully reading the text below and rating its politeness on a scale of 1 to 5, where ``1'' means ``not polite at all'' and ``5'' indicates ``very polite.''
    \item \textbf{Prompt 3}: Rate the politeness of the text below from 1 (not polite) to 5 (very polite).
\end{itemize}

\textbf{Wikipedia Toxicity Data}
\begin{itemize}
    \item \textbf{Prompt 1}: I want you to perform a data annotation task. Your task is to carefully read the text provided below and determine whether the text is toxic or not. In your output, I want you to respond with yes if the text is toxic, otherwise respond with no.
    \item \textbf{Prompt 2}: Perform data annotation by determining if the text is toxic or not. Only respond with yes or no.
    \item \textbf{Prompt 3}: Carefully review the text below and assess whether it is toxic. If the text is toxic, respond with yes; if not, respond with no.
\end{itemize}

\subsection{Extraction Prompts}

Previous studies have shown that prompting LLMs to be concise or prompting LLMs into a forced choice can reduce accuracy in response \cite{deng2024rephraserespondletlarge,deng2024rephraserespondletlarge}. For the POPQUORN Offensiveness and Politeness datasets, we used an extraction step to obtain the LLMs' ratings (from 1 to 5) from their responses to the classification prompts. Specifically, we used the following extraction prompt for the Offensiveness dataset:

\begin{quote}
Based on your response above, what is the numeric rating on the offensiveness scale? Only return a number with no other words, punctuation, or characters. If no number is provided, respond with ``N/A''.
\end{quote}

We used the following extraction prompt for the Politeness dataset:

\begin{quote}
Based on your response above, what is the numeric rating on the politeness scale? Only return a number with no other words, punctuation, or characters. If no number is provided, respond with ``N/A''.
\end{quote}

For the Wikipedia Toxicity dataset, we used the following extraction prompt: 

\begin{quote}
Based on your answer above, is the text toxic? Respond with only ``yes'' or ``no'' with no other words, punctuation, or characters.
\end{quote}

No extraction step was needed for the NLPositionality Toxicity dataset; all responses only needed light processing (e.g., removing \texttt{\textbackslash n}) to yield a ``yes'' or ``no'' answer.
\end{document}